\documentclass[11pt]{article}

\usepackage[final]{acl}

\usepackage{times}
\usepackage{latexsym}

\usepackage[T1]{fontenc}

\usepackage[utf8]{inputenc}

\usepackage{microtype}

\usepackage{inconsolata}

\usepackage{graphicx}

\usepackage{url}
\usepackage{mathtools}
\usepackage{adjustbox}
\usepackage{enumitem}
\usepackage{wrapfig}
\usepackage[table]{colortbl}
\usepackage{multirow}
\usepackage{makecell}
\usepackage[T1]{fontenc}
\usepackage{graphicx}
\usepackage{algorithm}
\usepackage{algpseudocode}
\usepackage[utf8]{inputenc}
\usepackage{babel}
\usepackage{amsmath}
\usepackage{amsthm}
\usepackage{amsfonts}
\usepackage[font=small]{caption}
\usepackage{tikz, tcolorbox}
\usepackage{pgfplots}
\usepackage{subcaption}
\usepackage{amsmath}
\usepackage{caption}
\usepackage{pgfplotstable}
\usepackage[linewidth=0.5pt]{mdframed}
\usepackage{tikz}
\usepackage{booktabs}
\usetikzlibrary{pgfplots.groupplots}
\usepackage{pifont}

\usepackage{soul}
\usepackage{listings}
\usepackage{courier}
\usepackage{xcolor}

\definecolor{hlblue}{RGB}{168,222,255}
\definecolor{hlyellow}{RGB}{255,245,168}
\definecolor{hlred}{RGB}{255,185,168}

\definecolor{myblue}{HTML}{268BD2}
\definecolor{mygreen}{HTML}{658354}
\definecolor{rebuttal}{rgb}{0,0,1}

\usepackage{minitoc}

\title{ModeX: Evaluator-Free Best-of-N Selection for Open-Ended Generation}

\author{Hyeong Kyu Choi \\
  University of Wisconsin-Madison \\
  \texttt{froilanchoi@cs.wisc.edu} \\\And
  Sharon Li~\thanks{Correspondence: \href{sharonli@cs.wisc.edu}{sharonli@cs.wisc.edu}}  \\
  University of Wisconsin-Madison \\
  \texttt{sharonli@cs.wisc.edu}\\
}

\begin{document}
\maketitle

\doparttoc 
\faketableofcontents 

\begin{abstract}
Selecting a single high-quality output from multiple stochastic generations remains a fundamental challenge for large language models~(LLMs), particularly in open-ended tasks where no canonical answer exists.
While Best-of-$N$ and self-consistency methods show that aggregating multiple generations can improve performance, existing approaches typically rely on external evaluators, reward models, or exact string-match voting, limiting their applicability and efficiency.
We propose \textbf{Mode Extraction}~(\textbf{ModeX}), an evaluator-free Best-of-$N$ selection framework that generalizes majority voting to open-ended text generation by identifying the \emph{modal} output representing the dominant semantic consensus among generated texts.
ModeX constructs a similarity graph over candidate generations and recursively applies spectral clustering to select a representative centroid, without requiring additional inference or auxiliary models.
We further instantiate this selection principle as \textbf{ModeX-Lite}, an improved version of ModeX with early pruning for efficiency.
Across open-ended tasks---including text summarization, code generation, and mathematical reasoning---our approaches consistently outperform standard single- and multi-path baselines, providing a computationally efficient solution for robust open-ended text generation.
Code is released in \url{https://github.com/deeplearning-wisc/ModeX}.
\end{abstract}
\section{Introduction}
\label{sec:intro}

Large language models~(LLMs) have demonstrated remarkable capabilities across a wide range of tasks, from code generation to creative writing~\cite{achiam2023gpt,team2023gemini,grattafiori2024llama,qwen2024qwen2,jiang2024mixtral}.
Despite this progress, reliably sampling a high-quality output from the model’s inherently stochastic generation process remains a fundamental challenge, particularly for open-ended tasks where no canonical answer exists.

Most LLM applications rely on \emph{single-path generation}, in which the model commits to a single output trajectory token by token.
This paradigm is inherently brittle: due to stochastic sampling, a single unfavorable token choice can trigger hallucinations or error propagation, even when the model's underlying distribution assigns substantial probability mass to correct or coherent outputs~\cite{wang2023self,wei2022chain}.
A natural solution is therefore to sample \emph{multiple} generation paths and select the best candidate.

\begin{figure}[t!]
    \centering
    \includegraphics[width=\linewidth]{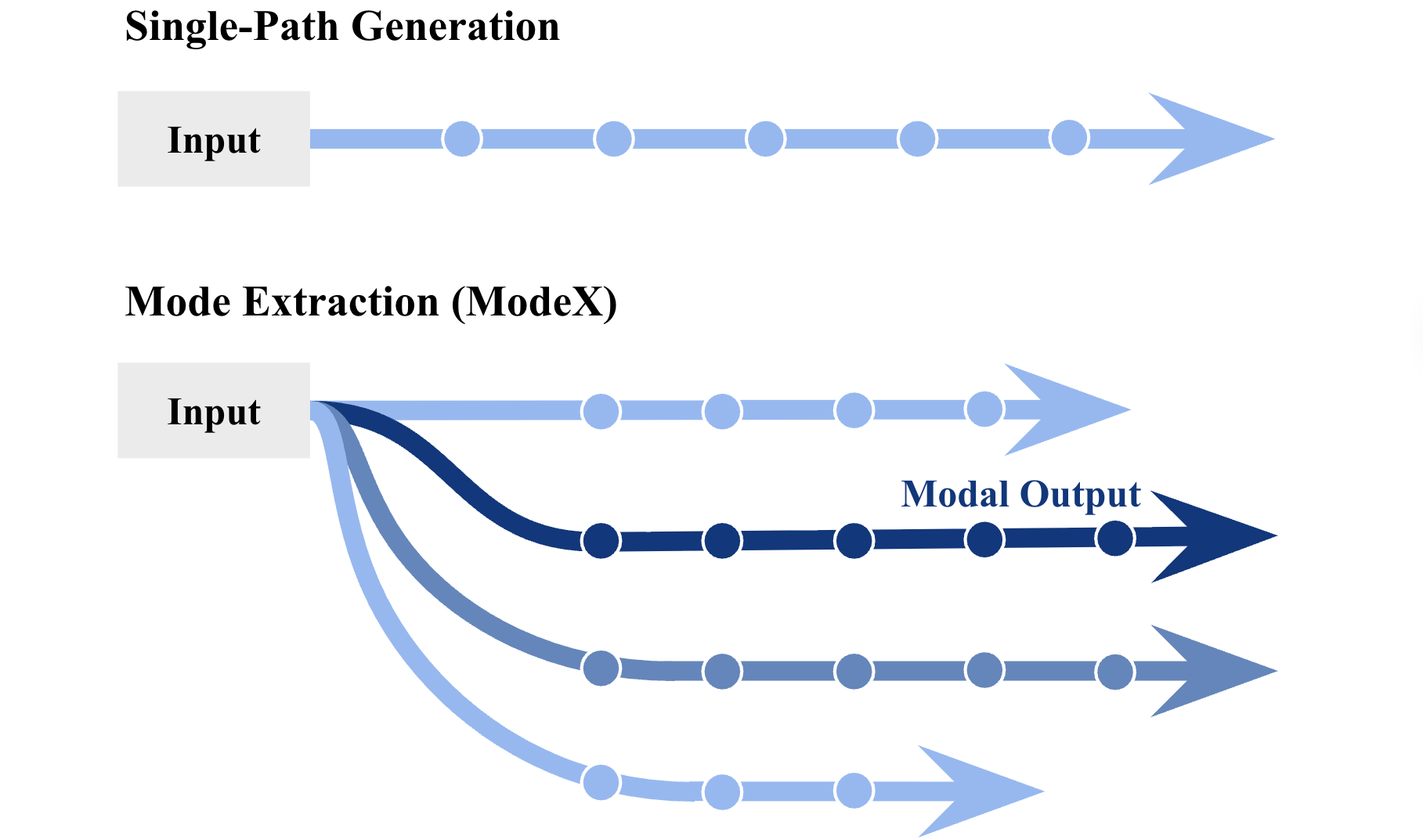}
    \caption{\textbf{Single Path Generation vs. Mode Extraction~(ModeX).} While single-path text generation commits to a single trajectory, ModeX leverages the structural information across multiple generation paths to select a ``modal'' output.}
    \label{fig:intro}
\end{figure}

Methods such as self-consistency and Best-of-$N$ sampling demonstrate that aggregating multiple outputs can substantially improve performance, particularly on reasoning tasks~\cite{wang2023self,hong2025slim,snell2024scaling}.
However, existing approaches typically rely on either (i) external evaluators such as reward models~\citep{cobbe2021training, lightman2023let} ~or (ii) exact string-match–based voting schemes.
Consequently, these methods are largely confined to closed-ended settings~(\textit{e.g.}, multiple-choice or short-answer tasks) and do not generalize naturally to open-ended text generation, where outputs may differ lexically yet remain semantically equivalent.
These limitations motivate a central question:
\textbf{\textit{Can we select a single high-quality output from multiple generation paths without external evaluators or significant computational overhead?}}

To address this question, we propose {Mode Extraction~(\textbf{ModeX})}, a  Best-of-$N$ selection framework that generalizes the principle of majority voting and self-consistency~\cite{wang2023self} to open-ended text generation.
Rather than relying on an external evaluator, ModeX operates directly within the {set of generated texts} to identify a representative, high-quality solution.
Concretely, ModeX builds a graph in which nodes correspond to generated sequences and edges encode pairwise lexical similarity.
We then apply spectral clustering---leveraging the Fiedler vector~\cite{fiedler1973algebraic} of the graph Laplacian---to isolate the dominant semantic cluster, and select its centroid as the final output.

Unlike standard voting schemes based on self-consistency, this procedure does not require exact string matches, predefined answer choices, or auxiliary scoring models.
Our key insight is that high-quality generations may vary lexically yet tend to form coherent clusters in the semantic space, whereas hallucinations and erroneous outputs are more likely to manifest as sparse outliers~\cite{lin2024generating}.
Consequently, the most reliable output is often not the most extreme or longest response, but the \emph{modal} one: the generation that best represents the dominant semantic consensus among samples~(Figure~\ref{fig:intro}).

Additionally, we show that the efficiency of ModeX can be further improved through early pruning of generation paths.
We introduce \textbf{ModeX-Lite}, a practical extension that periodically applies modal selection and pruning during generation.
By identifying non-representative trajectories at early stages, ModeX-Lite retains the robustness benefits of multi-path aggregation while incurring minimal computational overhead, enabling efficient and reliable generation in practice.
Through extensive experiments on text summarization, code generation, and mathematical reasoning, we demonstrate that our methods consistently outperform standard single- and multi-path baselines in both reliability and efficiency.
We summarize our contributions as follows:

\begin{enumerate}[leftmargin=*]
    \item We propose {ModeX}, an {evaluator-free} Best-of-$N$ selection framework that generalizes majority voting to open-ended generation, \textit{without requiring external evaluators or expensive computation}.
    \item We further introduce {ModeX--Lite}, an efficiency-improved variant that remains effective across a wide range of {any open-ended generation tasks.}
    \item We conduct extensive experiments on three open-ended generation tasks, showing {state-of-the-art performance among evaluator-free approaches}. We provide {theoretical justifications} of our approach, offering a principled Best-of-$N$ selection framework for modern LLMs.
\end{enumerate}

\begin{figure*}[t!]
    \centering
    \includegraphics[width=0.95\linewidth]{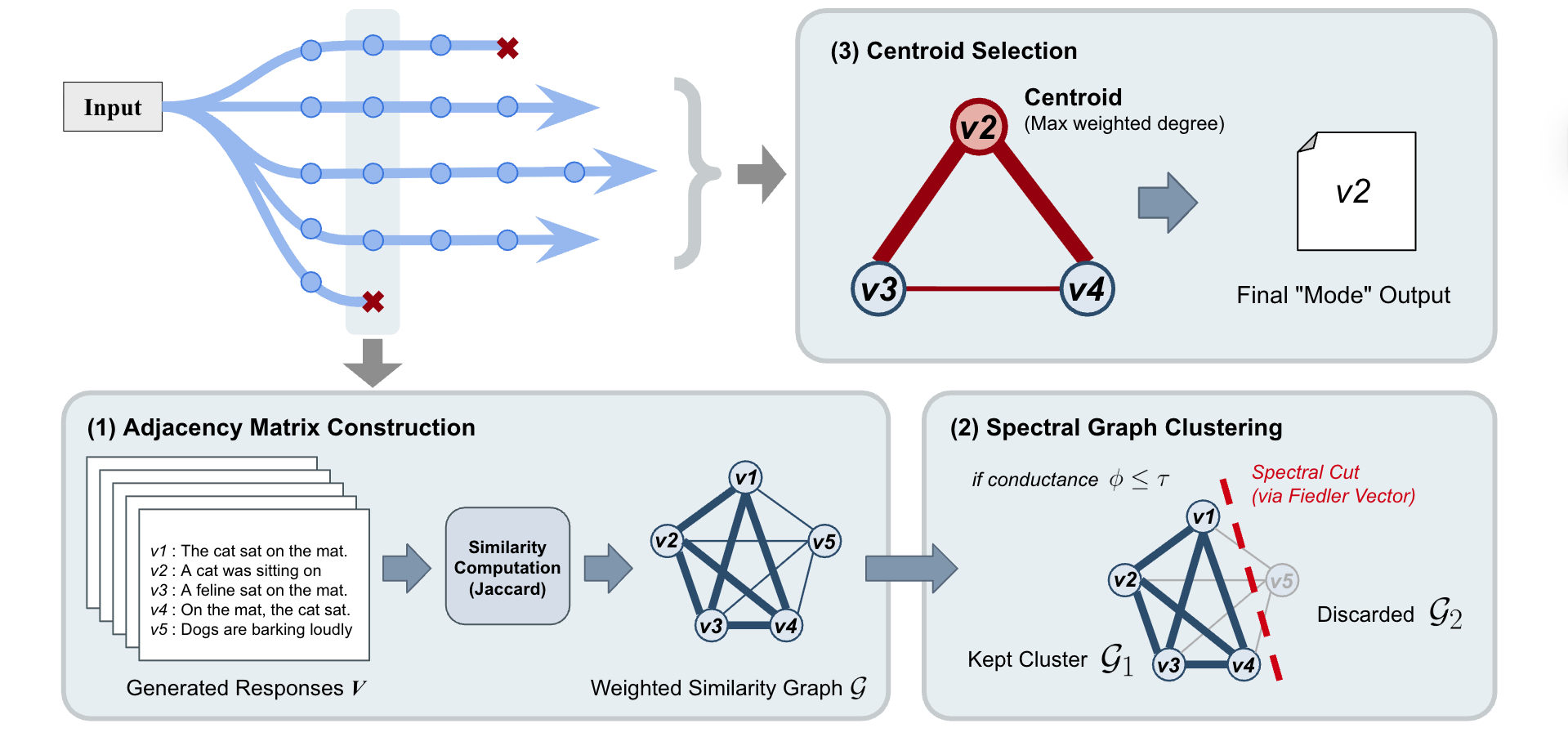}
    \caption{\textbf{Overview of the ModeX framework.} In standard ModeX, (1) adjacency matrix construction and (2) spectral graph clustering are iterated recursively as long as $\phi \leq \tau$. Then (3) centroid selection is performed. In the ModeX-Lite variant, (1) $\rightarrow$ (2) is performed only once without recursion for each pruning interval.} 
    \label{fig:mainfig}
\end{figure*}

\section{Discovering the Mode of Text}
\label{sec:avgoftext}
Can a single high-quality output be selected from multiple text generation paths without relying on reward models or external verifiers?
To address this question, we draw inspiration from the principles of \emph{majority voting} and \emph{self-consistency}, which have been widely adopted in multi-agent LLM frameworks for question answering~\cite{wang2023self,du2023improving,benedikt2025voting}.
These approaches rest on the premise that, as the number of sampled agents or generation trajectories increases, the aggregated response more faithfully reflects the underlying \emph{modal} belief of the LLM~\cite{choi2025debate,choi2026identity}.
In tasks with a finite answer space~(\textit{e.g.} multiple-choice question answering), simple voting schemes can therefore reliably recover the modal answer.

Extending this idea to open-ended text generation, however, introduces a fundamental challenge: when the output space becomes infinitely large, the notion of majority or mode is no longer directly countable. 
In this section, we tackle the problem of {identifying the modal generation in such open-ended tasks.}
We first introduce {Mode Extraction}~(ModeX), a graphical framework that enables principled mode approximation over multiple generated trajectories~(Section~\ref{sec:core}), and then qualitatively verify the effectiveness of this approach~(Section~\ref{sec:empirical_evidence}).

\subsection{Mode Extraction~(ModeX)}
\label{sec:core}

ModeX's approach to selecting the ``mode'' of the generated responses proceeds in three steps: (1) adjacency matrix construction, (2) graph spectral clustering, and (3) centroid selection.
A visual overview is provided in Figure~\ref{fig:mainfig}, and the corresponding pseudocode is presented in Algorithm~\ref{alg:mode_selection} of Appendix~\ref{apdx:algo}.

\paragraph{(1) Adjacency matrix construction.}
In closed-ended tasks~(\textit{e.g.}, multiple-choice question answering), majority voting can be viewed as the problem of identifying the largest cluster of identical responses.
This perspective naturally admits a graph-theoretic formulation.
Specifically, consider a graph $\mathcal{G} = (V, E)$, where each node $v \in V$ represents a generated response, and an edge $e \in E$ connects nodes that correspond to the same answer.
Under this construction, responses selecting the same choice form a clique, and the answer associated with the largest clique corresponds to the majority.
For instance, given five responses in which three select option ``A'' and two select ``B,'' the three ``A'' responses form the largest clique, and it is selected as the voted answer.

For open-ended generation, exact equivalence between responses is no longer well-defined, and the notion of a hard clique requires relaxation.
Thus, we define edges based on \emph{response similarity}.
Concretely, we construct a weighted adjacency matrix $A \in \mathbb{R}^{|V| \times |V|}$, where each entry measures the similarity between a pair of responses:
\begin{equation}
    A_{i,j} = s_1(v_i, v_j) + s_2(v_i, v_j) + s_3(v_i, v_j),
\end{equation}
with $v_i, v_j \in V$ denoting two generated responses.
Here, $s_1$, $s_2$, and $s_3$ correspond to Jaccard similarity computed over unigram, bigram, and trigram sets, respectively.
This construction yields a weighted graph where stronger edges indicate higher lexical overlap, allowing a soft generalization of voting to open-ended texts.
Comparison with an embedding similarity-based adjacency matrix is in Appendix~\ref{apdx:adjacency}.
Also, an ablation study on the n-gram components is provided in Appendix~\ref{apdx:ngram}.

\begin{figure*}[th!]
\begin{tcolorbox}[title={Chosen vs. Rejected}]
\begin{lstlisting}
(*\textbf{\color{blue}{[Chosen]}}*) Scientists propose that life on Earth originated around 3.8 billion years ago, with early DNA-like fragments, capable of (*{\sethlcolor{hlblue}\hl{self-assembly}}*), potentially guiding the evolution into complex life forms 4 billion years ago.Research from the University of Milan and the University of Colorado Boulder indicates (*{\sethlcolor{hlblue}\hl{RNA}}*)'s crucial role, as DNA fragments can naturally bond to form longer chains under certain conditions. This finding supports the idea that RNA could have acted as a template for early life, predating more complex molecules like DNA and proteins.

(*\textbf{\color{magenta}{[Rejected]}}*) Scientists theorize that life on Earth began evolving about (*{\sethlcolor{hlred}\hl{4 billion}*) years ago. Recent research by scientists from the University of Milan and University of Colorado Boulder suggests that DNA-like fragments present 4 billion years ago may have come with self-evolving 'instructions', leading to the formation of complex life. These fragments, potentially guided by their innate properties, evolved into (*\sethlcolor{hlyellow}\hl{longer chemical chains, possibly filling the gap between the simplest life forms and more advanced structures. The study supports the idea that these early DNA-like molecules could assemble and form longer chains}*) spontaneously under appropriate chemical conditions, paving the way for the development of life.

(*\textbf{\color{black}{[Target]}}*) Scientists say early DNA-like fragments guided their own growth. They claim the process can drive the formation of chemical bonds. These connect short DNA chains to form long ones for life to evolve. This (*{\sethlcolor{hlblue}\hl{self-assembly}*) capability has been shown to take place in (*{\sethlcolor{hlblue}\hl{RNA}*).
\end{lstlisting}
\end{tcolorbox}
\vspace{-3mm}
\caption{\textbf{Qualitative Examination.} In the text summarization task, ``rejected'' samples often \sethlcolor{hlblue}\hl{miss keywords}, \sethlcolor{hlred}\hl{include incorrect or less precise information}, and \sethlcolor{hlyellow}\hl{contain repetitive and verbose text}, whereas samples ``chosen'' by our method are overall concise.}
\label{fig:qual}
\end{figure*}

\paragraph{(2) Graph spectral clustering.}
To identify a dominant group of mutually consistent responses, we next perform clustering over the graph nodes.
A key challenge is that the number of coherent groups among generated responses is \emph{a priori} unknown.
Rather than fixing the number of clusters, we adopt a hierarchical spectral clustering approach that recursively partitions the graph into two subgraphs.

Specifically, given the weighted adjacency matrix $A$ and the corresponding degree matrix $D$, we compute the {Fiedler vector}~\cite{fiedler1973algebraic}, defined as the solution to the following problem:
\begin{equation}
    f = \underset{u^\top \mathbf{1} = 0, \|u\|_2 = 1}{\arg\min} u^\top (D - A) u,
\end{equation}
where $L = D - A$ denotes the graph Laplacian.
The Fiedler vector provides a continuous relaxation of the minimum cut objective and captures the most salient bipartition of the graph.
Further explanation is in Appendix~\ref{apdx:fiedler} for completeness.

We obtain a binary partition of the nodes by thresholding the entries of the Fiedler vector:
\begin{equation}
    c_i =
    \begin{cases}
        1, & \text{if } f_i \geq 0, \\
        0, & \text{otherwise},
    \end{cases}
\end{equation}
which induces a split of the vertex set $V = V_1 \cup V_2$.
To determine whether this partition corresponds to a meaningful separation, we evaluate the quality of the cut using the \emph{conductance ratio}~\cite{sinclair1992improved}.
The conductance of the resulting cut $(\mathcal{G}_1, \mathcal{G}_2)$ is:
\begin{equation}
    \phi(\mathcal{G}_1, \mathcal{G}_2)
    =
    \frac{\sum_{i \in V_1} \sum_{j \in V_2} w_{ij}}
    {\min\!\left(\sum_{i \in V_1} d_i,\; \sum_{i \in V_2} d_i\right)},
\label{eq:conductance}
\end{equation}
where $w_{ij}$ denotes the edge weight between nodes $i$ and $j$, and $d_i$ is the weighted degree of node $i$.
A lower conductance indicates a stronger separation between the two subgraphs.
Following the partition, we select the cluster containing the larger number of vertices; in the case of a tie, we select the cluster with the larger total edge weight.
We recursively apply this bipartitioning procedure until no further split yields a sufficiently low-conductance cut, \emph{i.e.}, when $\phi(\mathcal{G}_1, \mathcal{G}_2) \geq \tau$, at which point the recursion terminates.
In our experiments, we set the conductance threshold to $\tau = 0.8$ and analyze its effect in Section~\ref{sec:design_choice}.

\paragraph{(3) Centroid selection.}
Once the recursive spectral clustering procedure terminates, we obtain a final cluster of mutually consistent LLM outputs.
To extract a single representative response from this cluster, we select its \emph{centroid}, defined as the node that is most strongly connected to all other nodes in the cluster.
Formally, let $\tilde{A} \in \mathbb{R}^{n \times n}$ denote the adjacency matrix induced by the final cluster, where $n$ is the number of nodes in it.
We define the centroid as the node with the maximum weighted degree:
\begin{equation}
    v_c
    =
    \underset{{i \in \{1, \dots, n\}}}{\arg\max}
    \sum_{j=1}^{n} \tilde{A}_{ij}.
\end{equation}
Intuitively, this criterion selects the response that exhibits the highest overall similarity to other cluster members, and thus best represents the shared structure of the cluster.
The output corresponding to the selected centroid is interpreted as an approximation to the ``modal'' generation among the original set of $|V|$ sampled outputs.

\subsection{Qualitative Examination}
\label{sec:empirical_evidence}

To assess whether ModeX indeed selects a representative/modal output, we qualitatively compare the responses that are ultimately ``chosen'' with those that are not selected~(\textit{i.e.}, ``rejected'').
Figure~\ref{fig:qual} presents a representative example from the CNN/DailyMail text summarization benchmark~\cite{cnndaily1,cnndaily2}.
Across multiple samples, we observe that rejected summaries often omit important keywords, include imprecise or erroneous details, or exhibit repetitive and verbose phrasing.
These artifacts reflect idiosyncratic variations specific to individual generation paths and are less characteristic of an average response.
In contrast, the selected summaries are consistently concise and focused, capturing the key information of the source document.
These observations confirm that our approach is capable of identifying a representative output among multiple candidates, approximating the ``modal'' generation.
Additional qualitative examples are provided in Appendix~\ref{apdx:more_qual}, and theoretical discussions are in Section~\ref{sec:theory}.

\begin{figure}[t!]
    \centering
    \includegraphics[width=0.95\linewidth]{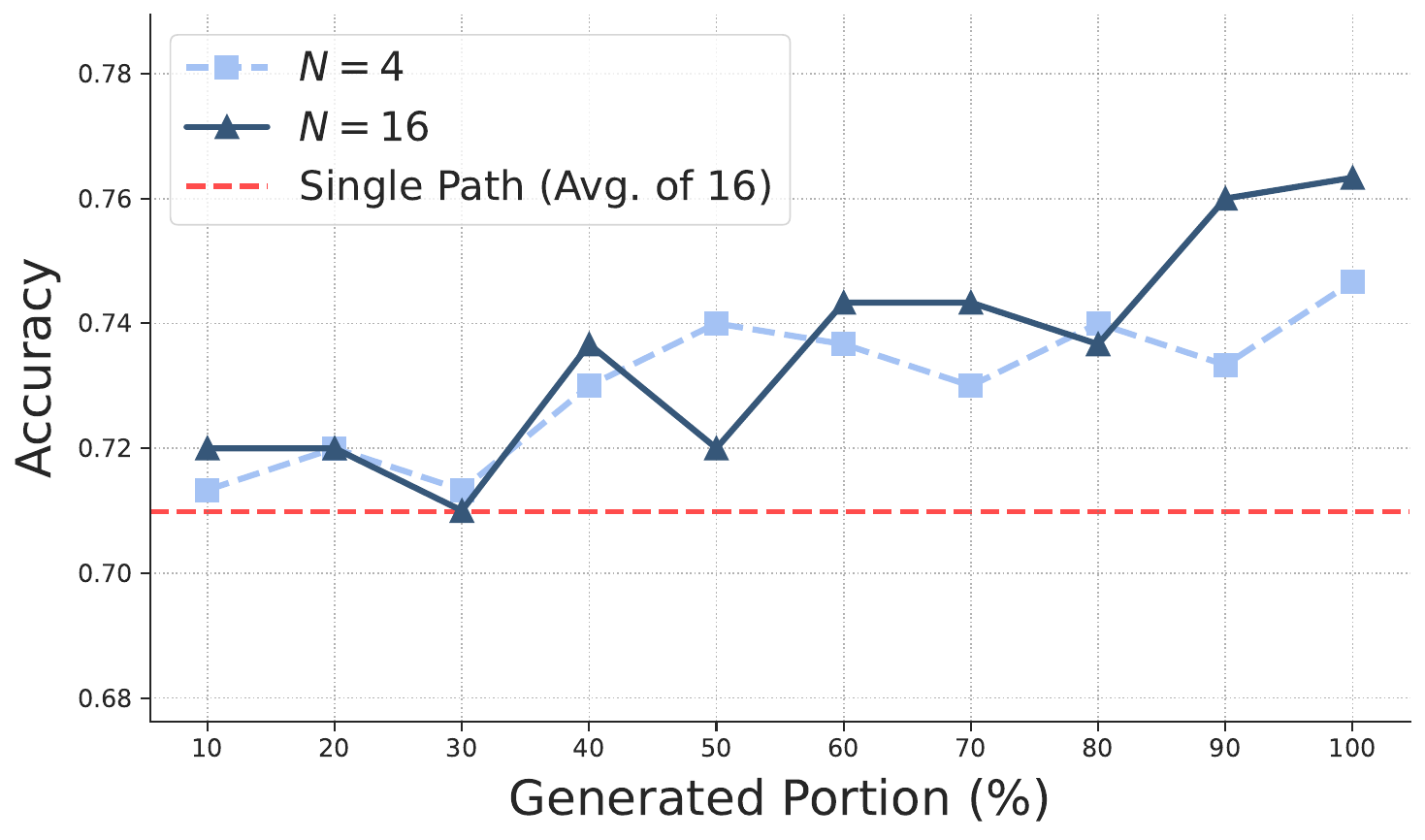}
    \caption{\textbf{Math reasoning accuracy at various stages of text generation.} Our mode selection approach consistently identifies high-quality samples early in the trajectory, maintaining high accuracy even with partial outputs.}
    \label{fig:early_stage}
\end{figure}

\section{Practical Extension: ModeX-Lite}
\label{sec:mupath}
Building on our principled selection framework that exploits the relational structure among multiple generated outputs, we now present a practical and computationally efficient extension.
Transformer-based architectures naturally support parallel sequence generation~\cite{vaswani2017attention}, enabling multiple generation trajectories concurrently.
This parallelism allows us to generate multiple candidate texts and identify a representative output among them without incurring substantial computational overhead.

\paragraph{Observation.}
Using our textual mode selection approach (Section~\ref{sec:avgoftext}), we observe that high-quality outputs can often be distinguished at early stages of the generation process.
As illustrated in Figure~\ref{fig:early_stage} for the math reasoning task with Qwen-7B, high-quality candidates are identifiable based on partial generations, \textit{even when less than 50\% of the full trajectory has been produced.}
This indicates that non-representative paths tend to diverge early, enabling them to be identified and pruned before generation is complete.

Motivated by this observation, we further introduce \textbf{ModeX-Lite}, a generation strategy that periodically prunes non-representative text paths at fixed intervals of $T$ steps~($T{=}100$ unless stated otherwise).
At each pruning interval, we apply graph spectral clustering to the partially generated trajectories, retaining only the most representative subset.
To ensure computational efficiency, spectral clustering is performed only once per pruning interval without recursion, and centroid selection is deferred until generation terminates.
This design balances the benefits of multi-path aggregation with practical computational efficiency.
For clarity, we illustrate the complete procedure in Algorithm~\ref{alg:mupath} and Figure~\ref{fig:mainfig}.

\begin{table*}[t!]
\caption{\textbf{Main results.} Performances of ModeX and ModeX-Lite on three task benchmarks---CNN/DailyMail~(text summarization), HumanEval~(code generation), and Math-500~(math reasoning)---are presented. Single Path reports the ``mean $\pm$ std" across 16 independent runs. Note that the code generation of Llama reports the performance of CodeLlama-7b-Instruct.}
\label{tbl:main}
\setlength{\tabcolsep}{5pt} 
\resizebox{\linewidth}{!}{%
\begin{tabular}{llcccc|cc|c}
\toprule
             &    & \multicolumn{4}{c}{\textbf{Text Summarization}}   & \multicolumn{2}{c}{\textbf{Code Generation}} & \textbf{Math Reasoning}    \\
\textbf{Model}  & \textbf{Method}           & {Rouge-1} & {Rouge-2} & {Rouge-L} & {BLEU} & {Pass@1}   & {BLEU}   & {Accuracy} \\
\midrule
Qwen     & Single Path      &    32.95 $\pm$ 0.36     &   10.47 $\pm$ 0.22     &  20.17 $\pm$ 0.28     &   3.37 $\pm$ 0.18    &   69.89 $\pm$ 3.59  &  7.92 $\pm$ 0.50    &   70.98 $\pm$ 1.74   \\
             & Self-refine      & 29.76   &  10.07  &  18.22  &   3.04    &  26.22   &   1.83              &  68.67   \\
             & LLM Judge~($N$=4)       &  32.91  &  10.54  &  20.09  &   3.19  &  70.12   &   7.23    &   71.67  \\
             & LLM Judge~($N$=16)       & 32.68   &  10.16  &  19.72  &  3.22  &   65.24  &   7.52    &   74.67  \\
             & Perplexity BoN~($N$=16)      & 34.28 & 11.24  & 21.06  & 3.92  & 73.17  &  8.18 & \textbf{78.00} \\
             & Self-Certainty BoN~($N$=16)   & 32.29 & 10.32  & 19.32  &  3.21 & 55.49  & 5.43 & 67.00 \\
             & \cellcolor{blue!6} ModeX ($N$=4)  &  \cellcolor{blue!6}  33.41  &   \cellcolor{blue!6} 10.81 &   \cellcolor{blue!6} 20.40  &    \cellcolor{blue!6} 3.53  &    \cellcolor{blue!6} 67.07  &  \cellcolor{blue!6} 8.02   &   \cellcolor{blue!6}  74.00 \\
             & \cellcolor{blue!6} ModeX ($N$=8)  &  \cellcolor{blue!6}  34.26  &   \cellcolor{blue!6} 11.39  &   \cellcolor{blue!6} 21.08 &    \cellcolor{blue!6}  3.59  &   \cellcolor{blue!6}  71.34  &  \cellcolor{blue!6}  \textbf{8.56} &   \cellcolor{blue!6} 74.67     \\
             & \cellcolor{blue!6} ModeX ($N$=16) &  \cellcolor{blue!6}  34.28  &   \cellcolor{blue!6}  11.24 &   \cellcolor{blue!6} 21.06   &    \cellcolor{blue!6} 3.92  &    \cellcolor{blue!6} 75.61   &  \cellcolor{blue!6} \underline{8.45}   &   \cellcolor{blue!6} \textbf{78.00}   \\
             & \cellcolor{blue!2} ModeX-Lite ($N$=4)  &  \cellcolor{blue!2} 34.15   &   \cellcolor{blue!2} 11.11 &   \cellcolor{blue!2}  21.13  &    \cellcolor{blue!2} 3.47  &    \cellcolor{blue!2} 73.17   &  \cellcolor{blue!2}  8.12   &   \cellcolor{blue!2} 72.67  \\
             & \cellcolor{blue!2} ModeX-Lite ($N$=8)  &  \cellcolor{blue!2} \underline{35.21}   &   \cellcolor{blue!2} \underline{12.04} &   \cellcolor{blue!2} \underline{21.83} &    \cellcolor{blue!2} \underline{4.05}  &   \cellcolor{blue!2}  \underline{76.22}  &  \cellcolor{blue!2}  8.42  &   \cellcolor{blue!2} 74.67     \\
             & \cellcolor{blue!2} ModeX-Lite ($N$=16) &  \cellcolor{blue!2} \textbf{35.78}   &   \cellcolor{blue!2} \textbf{12.35}  &   \cellcolor{blue!2}  \textbf{21.89}  &    \cellcolor{blue!2} \textbf{4.36}  &    \cellcolor{blue!2}  \textbf{78.66}  &  \cellcolor{blue!2}  8.29   &   \cellcolor{blue!2}  \underline{75.33} \\
             & \cellcolor{olive!6} Best-of-$16$~(Gold Standard) &  \cellcolor{olive!6}  33.46  &   \cellcolor{olive!6} 10.64  &   \cellcolor{olive!6}  20.49  &    \cellcolor{olive!6}  3.26  &    \cellcolor{olive!6} --  &  \cellcolor{olive!6}  --   &   \cellcolor{olive!6} 82.00   \\ 
\midrule
Llama     & Single Path      &    33.97 $\pm$ 0.49     &   12.15 $\pm$ 0.22    &    21.30 $\pm$ 0.34    & 4.41 $\pm$ 0.17     &   18.29 $\pm$ 15.22     &   4.94 $\pm$ 1.97    &     38.75 $\pm$ 1.98     \\
             & Self-refine      &  23.97  &  8.83  &  15.28  &  2.75   &  3.05   &   1.71              &  39.00   \\
             & LLM Judge~($N$=4)       &  34.33  &  12.55  & 21.48   &  4.62   &   12.80  &  3.72   &  37.33   \\
                 & LLM Judge~($N$=16)       &  34.54  &  12.57  &  21.60  &  4.67  &  7.32   &   3.14  &  38.67   \\
             & Perplexity BoN~($N$=16)      & 34.41 &  12.45 & 21.88  &  4.73 &  \textbf{33.54} & 5.81  & \underline{48.00} \\
             & Self-Certainty BoN~($N$=16)   & 32.42 & 11.77  & 20.06  &  4.12 & 4.27  & 1.37  & 27.33 \\
             & \cellcolor{blue!6} ModeX ($N$=4)  &  \cellcolor{blue!6}  35.01  &   \cellcolor{blue!6} 12.75 &   \cellcolor{blue!6} 22.04  &    \cellcolor{blue!6} 4.75  &    \cellcolor{blue!6} 23.78  &  \cellcolor{blue!6}  6.20  &   \cellcolor{blue!6}  43.00 \\
             & \cellcolor{blue!6} ModeX ($N$=8)  &  \cellcolor{blue!6}  35.26  &   \cellcolor{blue!6} 12.97 &   \cellcolor{blue!6} 22.13  &    \cellcolor{blue!6} 4.65  &   \cellcolor{blue!6}  27.44  &  \cellcolor{blue!6} 6.39  &   \cellcolor{blue!6}   43.67   \\
             & \cellcolor{blue!6} ModeX ($N$=16) &  \cellcolor{blue!6}  \textbf{35.79}  &   \cellcolor{blue!6}  \textbf{13.35} &   \cellcolor{blue!6} \underline{22.70}   &    \cellcolor{blue!6}  \underline{5.13} &    \cellcolor{blue!6} \underline{32.32}   &  \cellcolor{blue!6} \underline{7.35}   &   \cellcolor{blue!6}  \textbf{49.33}  \\
             & \cellcolor{blue!2} ModeX-Lite ($N$=4)  &  \cellcolor{blue!2}  35.28  &   \cellcolor{blue!2} 12.87 &   \cellcolor{blue!2}  22.02  &    \cellcolor{blue!2} 4.63  &   \cellcolor{blue!2}  20.12  &  \cellcolor{blue!2} 5.65 &   \cellcolor{blue!2} 39.00     \\
             & \cellcolor{blue!2} ModeX-Lite ($N$=8)  &  \cellcolor{blue!2} 34.46   &   \cellcolor{blue!2} 12.60 &   \cellcolor{blue!2}  21.77  &    \cellcolor{blue!2} 4.40  &    \cellcolor{blue!2}  26.22  &  \cellcolor{blue!2}  6.56   &   \cellcolor{blue!2}  42.33    \\
             & \cellcolor{blue!2} ModeX-Lite ($N$=16) &  \cellcolor{blue!2} \underline{35.57}   &   \cellcolor{blue!2} \underline{13.22} &   \cellcolor{blue!2}  \textbf{22.80}  &    \cellcolor{blue!2} \textbf{5.26}  &  \cellcolor{blue!2}  {29.88}  &  \cellcolor{blue!2}  \textbf{7.77}   &   \cellcolor{blue!2}  {45.33}    \\ 
             & \cellcolor{olive!6} Best-of-$16$~(Gold Standard) &  \cellcolor{olive!6} 35.68   &   \cellcolor{olive!6} 13.02  &   \cellcolor{olive!6}  22.25  &    \cellcolor{olive!6}  4.90  &    \cellcolor{olive!6} --  &  \cellcolor{olive!6}  --   &   \cellcolor{olive!6}  63.00  \\
\bottomrule
\end{tabular}
}
\end{table*}

\section{Experiments}

\subsection{Setup}

\paragraph{Tasks and Models.} 
We test on three representative open-ended tasks: text summarization with CNN/DailyMail~\cite{cnndaily1,cnndaily2}, code generation with HumanEval~\cite{chen2021evaluating}, and mathematical reasoning with Math-500~\cite{lightman2023lets}. 
Details on tasks, models, reward models, and metrics are in Appendix~\ref{apdx:details} due to limited space.

\vspace{-2mm}
\paragraph{Baselines.} 
We compare our method against four baselines: 
(1) \textit{Single Path} reports the performance of standard single-path generation, averaged across 16 independent runs; 
(2) \textit{Self Refine}~\cite{madaan2023self} iteratively modifies an output four times, as performance is typically known to saturate by this point; 
(3) \textit{LLM Judge}~\cite{zheng2023judging} employs a separate LLM to select the best output out of either 4 or 16 candidates;
(4) Perplexity selects the output with the lowest average uncertainty;
(5) Self-Certainty~\cite{kang2025scalable} chooses the output with the lowest negative log likelihood;
(6) \textit{Best-of-N} serves as the gold-standard reference, utilizing reward models to choose the best among $N=16$ samples.
Prompt templates are in Appendix~\ref{apdx:prompts}.

\subsection{Experimental Results}
\label{sec:exp_results}
\paragraph{ModeX consistently outperforms baselines.} 
As shown in Table~\ref{tbl:main}, our method achieves consistently strong performance across all evaluated datasets.
In particular, applying ModeX to Qwen with $N{=}16$ generation paths improves the mean \textit{Single-Path} baseline from 69.89\% to \textbf{78.66\%} on the code generation task Pass@1 metric.
Moreover, ModeX outperforms LLM Judge with 16 candidates by significant margins, and even sometimes surpasses the gold standard Best-of-$N$ that requires external evaluators.
This demonstrates that our evaluator-free selection mechanism is more effective than approaches that rely on an LLM to rank or verify multiple outputs.
When comparing with the latest approach self-certainty~\citep{kang2025scalable}, ModeX shows generally superior performance across tasks.
Overall, these results indicate that ModeX effectively harnesses the benefits of ensemble generation, yielding substantial gains without introducing additional supervision.

\paragraph{More compute does not surpass Single Path performance.} 
Despite consuming roughly $4\times$ the computational resources of standard text generation, \textit{Self Refine} fails to surpass our ModeX approaches.
In fact, we observe that the refinement process can cause performance to drop significantly below the original Single Path baseline. 
This suggests that simply scaling inference compute via self-correction is not effective without a selection mechanism to filter out error propagation, compared to parallel text generation.

\begin{figure*}[t!]
    \centering
    \includegraphics[width=0.9\linewidth]{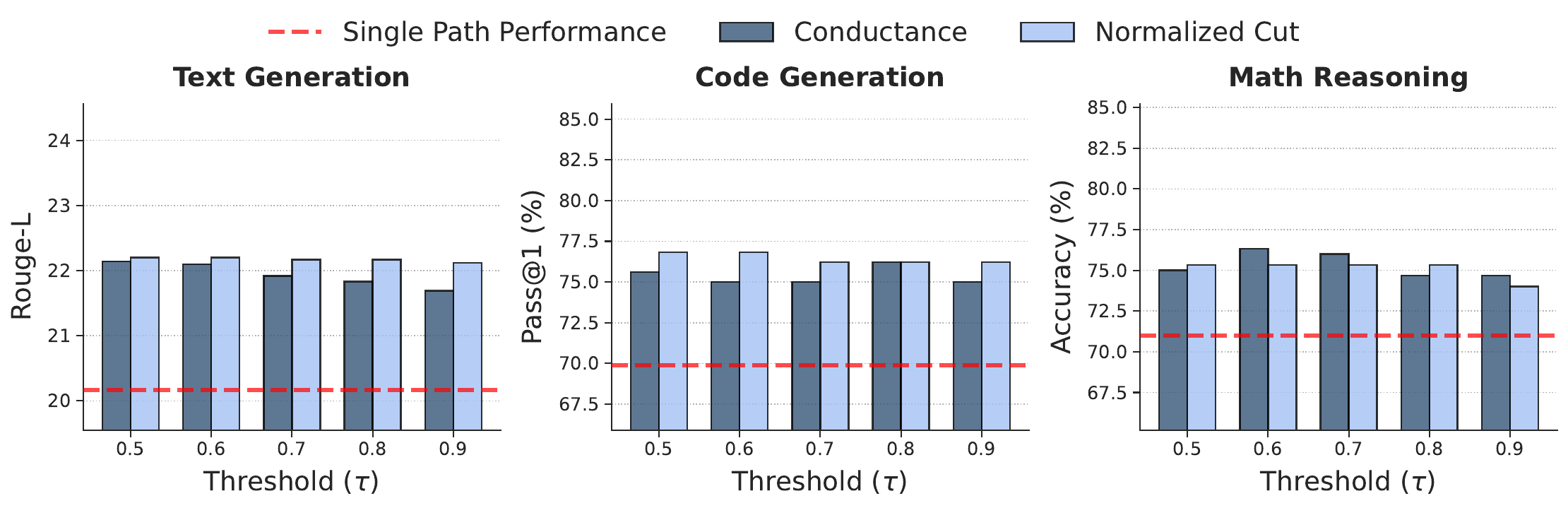}
    \includegraphics[width=0.9\linewidth]{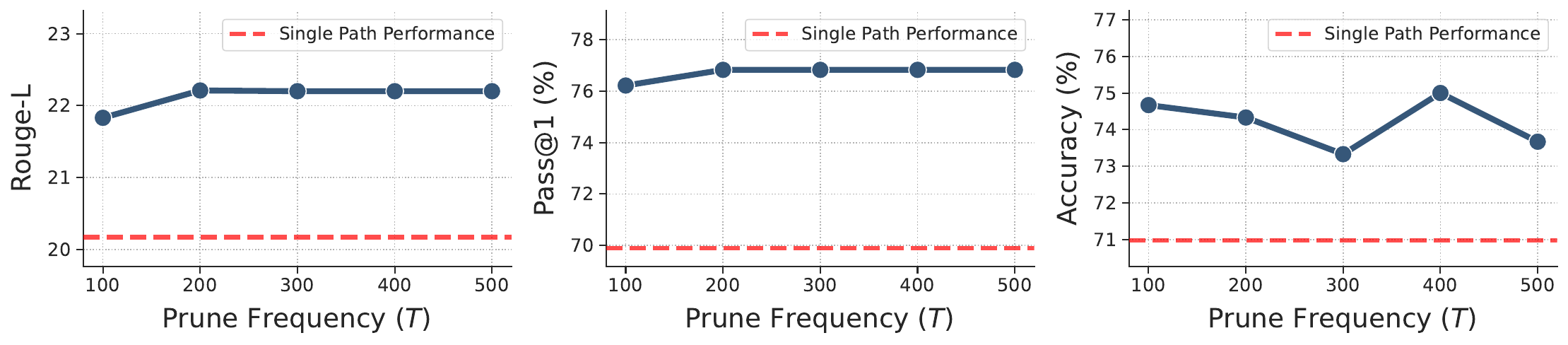}
    \caption{\textbf{Sensitivity analysis.} ModeX-Lite shows performance consistently above the single-path baseline in all settings.}
    \label{fig:hyperparam}
\end{figure*}
\paragraph{Increasing the number of generation paths generally improves performance.}
We further investigate the effect of the number of generation paths, $N$, on overall performance, as summarized in Table~\ref{tbl:main}.
While both the \textit{LLM Judge} baseline and ModeX variants can in principle benefit from the larger search space induced by additional generation paths, ModeX exhibits substantially more consistent and scalable gains as $N$ increases.
In the math reasoning task with Llama, increasing the number of paths from $N{=}4$ to $N{=}16$ yields only a marginal +1.34 percentage-point improvement in accuracy for the \textit{LLM Judge} baseline.
In contrast, ModeX-Lite leverages the same increase in paths to achieve a +7.33 percentage-point gain.
These results indicate that merely generating more candidates is insufficient; instead, a principled, structure-aware selection strategy is essential to effectively exploit the diversity of the generation space.

\section{Discussions}
In this section, we provide a deeper analysis of ModeX and ModeX-Lite. 
We first examine the impact of key design choices and hyperparameter sensitivity~(Section~\ref{sec:design_choice}), and provide a complexity analysis demonstrating the computational efficiency of our approach~(Section~\ref{sec:complexity}).
We also formalize the theoretical connection between our graph-based selection mechanism and modal approximation~(Section~\ref{sec:theory}).

\subsection{Sensitivity Analysis}
\label{sec:design_choice}
We analyze the sensitivity of ModeX-Lite to three key design choices: (a) graph partitioning objective, (b) spectral threshold $\tau$ (Eq.~\eqref{eq:conductance}), and (c) pruning frequency $T$.
In the top panel of Figure~\ref{fig:hyperparam}, we benchmark our conductance criterion~(varying $\tau \in \{0.5, \dots, 0.8\}$) against an alternative, Normalized Cut~\cite{shi2000normalized}:
\begin{equation*}
    \phi'(\mathcal{G}_1, \mathcal{G}_2)
    = \sum_{i \in V_1} \sum_{j \in V_2} w_{ij} \bigg(\frac{1}{\sum_{i \in V_1} d_i} + \frac{1}{\sum_{i \in V_2} d_i}\bigg),
\end{equation*}
where $V_1$ and $V_2$ are the set of nodes in subgraphs $\mathcal{G}_1$ and $\mathcal{G}_2$.
In the bottom panel, we examine the impact of the pruning frequency $T \in \{100, \dots, 500\}$.
Overall, we observe that performance is remarkably robust to hyperparameter variations; our method shows relatively stable performance across design choices, and consistently yields significant improvements over the single-path baseline~(red dashed line) across all tested configurations.

\begin{table}[t!]
\caption{\textbf{Complexity and Latency Analysis.} $L$: Sequence length~(\# of tokens), $N$: Number of paths, $k$: Refinement iterations, $L_{judge}$: Length of judge output. We assume parallel generation for $N > 1$. Latency reports the per sample wall time measured on CNN/DailyMail with Qwen-7B.}
\label{tbl:complexity}
\resizebox{\linewidth}{!}{%
\setlength{\tabcolsep}{10pt} 
\begin{tabular}{l|cc}
\toprule
\textbf{Method} & \textbf{Complexity} & \textbf{Latency ($s$)}  \\
\midrule
Single Path & $\mathcal{O}(L)$ & 5.5 \\
Self-Refine & $\mathcal{O}(kL)$ & 31.7 \\
LLM Judge & $\mathcal{O}(NL + N L_{judge})$ & 10.7 \\
Best-of-$N$ & $\mathcal{O}(NL + N \cdot C_\text{RM})$ &  11.1 \\
\textbf{ModeX-Lite ($N=4$)} & $\mathcal{O}(NL + N^2)$ & 7.2 \\
\textbf{ModeX-Lite ($N=16$)} & $\mathcal{O}(NL + N^2)$ & 9.1 \\
\bottomrule
\end{tabular}
}
\end{table}
\subsection{Complexity Analysis}
\label{sec:complexity}
We assess the efficiency of ModeX-Lite by comparing its computational complexity and empirical latency against standard baselines~(Table~\ref{tbl:complexity}). 
While single-path generation scales linearly with sequence length~($\mathcal{O}(L)$), baseline strategies often introduce significant overhead: \textit{Self-Refine} suffers from sequential dependency~($\mathcal{O}(kL)$), and \textit{LLM Judge} requires a computationally expensive second inference pass, and Best-of-$N$ may require auxiliary reward model passes ($\mathcal{O}(C_{RM})$) which can be expensive in real-world applications without ground-truth labels to evaluate the outputs.
In contrast, ModeX's complexity is dominated by the parallel generation of $N$ trajectories~($\mathcal{O}(NL)$). 
The subsequent selection step---spectral clustering---scales as $\mathcal{O}(N^2)$, which is negligible in practice~($N \ll L$) and requires no neural re-evaluation.
Empirically, this architectural difference translates into substantial latency gains. As shown in Table~\ref{tbl:complexity}, \textit{Self-Refine} incurs a massive latency penalty ($31.7$s) due to its serial nature. 
ModeX ($N=16$) achieves a $\mathbf{3.5\times}$ speedup ($9.1$s) over this baseline while maintaining robust performance. 
Also, compared to the \textit{LLM Judge} ($10.7$s), our method is faster because its selection mechanism is ``evaluator-free", deriving the optimal path solely from the relational structure of texts. 
With $N=4$, ModeX-Lite ($7.2$s) adds only minor overhead to the \textit{Single Path} baseline ($5.5$s).

\subsection{Theoretical Analysis}
\label{sec:theory}
To formally justify ModeX's graph-based selection mechanism, we model the text generation process as sampling from a high-dimensional probability distribution. 
We show that under mild assumptions, our two-step process, spectral clustering $\rightarrow$ centroid selection, corresponds to identifying the \textit{modal region} of the distribution and then estimating the \textit{mode}~(peak density) within that region.

\paragraph{Setup.}
Let $\mathcal{X}$ be the space of all possible generated texts. Let $p(x)$ be the probability density function defined over $\mathcal{X}$ by the LLM given a specific prompt. 
We observe a set of $N$ i.i.d. samples $V = \{v_1, v_2, \dots, v_N\}$ drawn from $p(x)$.
Our goal is to identify the sample $v^* \in V$ that is closest to the true mode of the distribution:
\begin{equation}
    v^* \approx \operatorname*{arg\,max}_{x \in \mathcal{X}} p(x)
\end{equation}
Our approach rests on the hypothesis that the generation process draws samples from a potentially multi-modal distribution $p(x)$.
For instance, in multiple-choice tasks, distinct modes typically emerge around competing options like `A' and `B'.
We therefore address mode identification in two steps: first, isolating a coherent, high-density region~(via spectral clustering), and second, estimating the point of maximum density within that region~(via degree centrality).

\paragraph{Theorem 1. (Spectral Clustering Isolates Modal Components)}
\textit{Consider a distribution $p(x)$ supported on a disjoint union of manifolds $\mathcal{M}_1 \cup \mathcal{M}_2$ (representing distinct semantic modes) separated by a region of low density. As $N \to \infty$, the spectral bipartition based on the Fiedler vector converges to the geometric cut that separates $\mathcal{M}_1$ and $\mathcal{M}_2$ with minimum probability flow.}

\noindent \textit{Proof.} See Appendix~\ref{apdx:proofs}.
\vspace{2mm}

While the Fiedler vector produces a binary partition, our recursive framework naturally generalizes to distributions with $K > 2$ modes. 
We view the clustering as a hierarchical decomposition of the probability space: each spectral cut splits the current set of samples into two disjoint sets of semantic manifolds. 
By recursively applying this bipartition until the conductance criterion is met, we effectively isolate a single dominant mode from the original mixture of $K$ modes.

Once the recursive spectral clustering terminates, we obtain a subgraph of $V' \subseteq V$ assumed to be drawn from a locally uni-modal component of the distribution. 
We now show that the degree centrality within this cluster identifies the mode:
\paragraph{Theorem 2. (Weighted Degree as KDE)}
\textit{Given a set of samples $V'$ drawn from a distribution, let $K: \mathcal{X} \times \mathcal{X} \to \mathbb{R}_{\ge 0}$ be a symmetric similarity kernel~(e.g., cosine or Jaccard similarity). The weighted degree $d(v_i) = \sum_{v_j \in V'} K(v_i, v_j)$ is proportional to the Kernel Density Estimator (KDE) of the underlying probability density $p(x)$.}

\noindent \textit{Proof.} See Appendix~\ref{apdx:proofs}.
\vspace{2mm}

Consequently, our two-step process performs a \textit{conditional mode estimation}: by first partitioning the graph to isolate the dominant cluster $C$~(Theorem 1), the subsequent centroid selection identifies the sample $x = v^*$ that maximizes the conditional likelihood $p(x \mid x \in C)$~(Theorem 2), thereby recovering the specific mode of the dominant interpretation.
In effect, this replaces the discrete frequency counting of exact matches in majority voting with continuous density estimation over semantic manifolds.
\textit{\textbf{This framework therefore constitutes a formal generalization of ``majority voting'' to open-ended generation tasks.}}

\section{Related Works}
\vspace{-1mm}
\paragraph{LLM Generation Strategy.}
A growing line of work has proposed enhanced generation strategies that go beyond standard single-path generation.
One approach incorporates \emph{reward models or external verifiers} at inference time to guide generation toward preferred outputs~\cite{huang2025deal,hung2025reward,khanov2024args,deng2023reward,ouyang2022training}.
Another line of research exploits \emph{internal model signals} from internal representations and output embeddings~\cite{manvi2024adaptive,chuang2023dola}.
More recently, \emph{multi-agent generation} frameworks have been introduced, in which multiple agents or experts collaborate during generation by alternately proposing tokens to produce a single output stream~\cite{chakraborty2025collab}.
While effective, these approaches focus on refining a \emph{single} generation path and often require additional models and coordination mechanisms.

\paragraph{Multi-Path Text Generation.}
A promising avenue for enhancing generation quality involves explicitly leveraging \emph{multiple} generation trajectories.
The standard approach, \emph{Best-of-$N$}~(BoN), samples independent candidates and selects the optimal output via an external reward model~\cite{sun2024fast,ouyang2022training}. 
While effective, BoN may incur high computational costs and relies heavily on the quality of the external evaluator. 
Alternative strategies have attempted to mitigate this via the notion of self-consistency~\cite{hong2025slim,yin2024aggregation,wang2023self,chen2025we}, internal model signals~\cite{liang2025clue,kang2025scalable}, external reward models~\cite{snell2024scaling}, or multi-agent collaboration~\cite{wang2025mixture,chakraborty2025collab}. 
Yet, most methods typically target exact-match answer aggregation, restricting their utility to closed-ended reasoning tasks. 
In this work, we bridge this gap by introducing a framework \textit{applicable to any open-ended tasks} that functions \textit{without external evaluators}.

\section{Conclusion}

We introduced ModeX and ModeX-Lite, a principled framework that generalizes majority voting to open-ended text generation by identifying modal outputs through graph-based clustering over multiple sampled trajectories. 
Our approach enables effective Best-of-N selection without external evaluators, consistently outperforming strong baselines across summarization, code generation, and mathematical reasoning while remaining computationally efficient. 
Beyond empirical gains, ModeX highlights that improvements from multi-path generation critically depend on structured aggregation rather than increased sampling alone, offering a new perspective on inference-time scaling by leveraging the intrinsic distributional structure of model outputs. 
This work suggests that reliable generation can emerge from internal consensus, and points to future directions such as richer similarity measures and adaptive generation–selection strategies. for further improving robustness and efficiency.

\section*{Limitations}
While ModeX offers a robust, inference-only selection mechanism, it relies on lexical Jaccard similarity to approximate semantic consensus; this metric may fail to recognize valid paraphrases that differ significantly in surface form, potentially causing the rejection of high-quality but lexically distinct outputs.
Further investigation with embedding-based similarity measures may be useful.
Moreover, the method rests on the assumption that the most frequent output is correct; in cases where the underlying model exhibits systematic bias or ``mode collapse" towards a specific hallucination, our spectral clustering approach may inadvertently identify and reinforce this consensus on error.
However, we view this as a systematic error of the target LLM itself, rather than a direct limitation of ModeX.
Relevant future work to mitigate such corner cases is called for.

\section*{Ethical Considerations}
This work aims to improve the consistency and reliability of LLMs without relying on costly external verification. We acknowledge that adopting multi-path generation strategies increases the aggregate energy consumption per query, contributing to a larger environmental footprint. 
We affirm that our experiments utilize public benchmarks and do not involve human subjects, and while improved reasoning capabilities could theoretically be misused, our focus remains on mitigating hallucinations and enhancing general model robustness. 

\section*{Disclosure of LLM Usage} 
We used large language model~(LLM) tools to polish portions of the writing, to assist in literature searches to check for relevant related work that we might have missed, and to check sanity of our theoretical claims.

\section*{Acknowledgement}
The authors would like to thank Min-Hsuan Yeh and Wendi Li for their valuable comments on
the manuscript. Hyeong Kyu Choi and Sharon Li are supported in part by the AFOSR Young
Investigator Program under award number FA9550-23-1-0184, National Science Foundation under
awards IIS-2237037 and IIS-2331669, Office of Naval Research under grant number N00014-23-
1-2643, Schmidt Sciences Foundation, Open Philanthropy, Alfred P. Sloan Fellowship, and gifts
from Google and Amazon.

\newpage


\bibliography{custom}

\clearpage

\onecolumn\appendix

\addcontentsline{toc}{section}{Appendix}
\part{Appendix} 
\parttoc 

\twocolumn
\section{Experimental Details}
\label{apdx:details}

\subsection{Benchmark Details}

\paragraph{Text Summarization.}
For the text summarization task, we evaluate on the CNN/DailyMail benchmark~\cite{cnndaily1,cnndaily2}, which is a dataset for abstractive text summarization. It was constructed from news articles from CNN and Daily Mail. We utilize the first 300 samples from the test split of version 3.0.0.

\paragraph{Code Generation.}
For the code generation task, we evaluate on the HumanEval benchmark~\cite{chen2021evaluating}, which contains 164 Python programming problems with a function signature, docstring, body, and several unit tests. We utilize the full dataset of the test split.

\paragraph{Mathematical Reasoning.} 
For the mathematical reasoning task, we evaluate on the Math-500 benchmark~\cite{lightman2023lets}, which contains 500 math questions, ranging six domains, including algebra, geometry, intermediate algebra, number theory, precalculus, and probability.
We utilize the first 300 samples from the test split. 
Also, for precise evaluation, we adopt the evaluation protocol from \citet{sheng2024hybridflow}'s codebase.

\subsection{Model Details}
we evaluate on two model families: \texttt{Qwen2.5-7b-instruct}~\cite{yang2024qwen2} and \texttt{Llama3.1-8b-instruct}~\cite{grattafiori2024llama}. 
For the code generation task, we adopt \texttt{CodeLlama-7b-Instruct}~\cite{roziere2023code}, instead of \texttt{Llama3.1-8b-instruct}.
For the gold-standard Best-of-$N$, we adopt the \texttt{Skywork-Reward-V2-Qwen3-8B}~\cite{liu2025skywork} reward model for the text summarization task evaluation, and \texttt{Qwen2.5-Math-PRM-7B}~\cite{prmlessons} for the math reasoning task.
The code generation task does not currently have a good reward model for Best-of-$N$ selection.

\subsection{Metric Details}

\paragraph{ROUGE-1} is a recall-oriented metric that measures the overlap of unigrams~(individual words) between the generated text and a reference text. It assesses how much of the key content from the reference appears in the output.

\paragraph{ROUGE-2} is similar to ROUGE-1, but measures the overlap of bigrams~(pairs of consecutive words). This captures some level of fluency and phrasing, rather than just isolated keywords.

\paragraph{ROUGE-L} is based on the Longest Common Subsequence~(LCS) between the generated text and the reference. Unlike ROUGE-1 or ROUGE-2, it does not require a fixed n-gram length. Instead, it identifies the longest sequence of words that appear in both texts in the same relative order~(though not necessarily consecutively). This allows it to capture sentence-level structure and flow better than simple keyword matching.

\paragraph{BLEU} is a precision-oriented metric that counts the overlap of n-grams~(usually 1 to 4) between the generation and the reference, penalizing outputs that are too short~(brevity penalty). It is widely used to assess how ``natural" or close to a human reference the generation is. BLEU is adopted for both text summarization and code generation tasks, but the importance of this metric is lower for the latter task.

\paragraph{Pass@1} is a functional correctness metric often used in code generation or math reasoning. It measures the percentage of problems where the model's first single attempt is correct (\textit{i.e.}, passes all unit tests or yields the correct final answer).

\section{Prompt Templates}
\label{apdx:prompts}

\subsection{Task Prompts}
\vspace{2mm}

\begin{figure}[h!]
    \centering
    \begin{tcolorbox}[title={Text Summarization}]\footnotesize

Summarize the following in less than 5 sentences:

\texttt{<TEXT TO SUMMARIZE>}

First, briefly state your step-by-step reasoning. Then, make sure to provide your summary after stating ``\# Answer \#"

    \end{tcolorbox} 
\end{figure}

\begin{figure}[h!]
    \centering
    \begin{tcolorbox}[title={Code Generation}]\footnotesize
Complete the following code:

\texttt{<BEGINNING OF CODE>}

First, briefly state your step-by-step reasoning. Then, make sure to provide ONLY your COMPLETE code after stating ``\# Code \#"

    \end{tcolorbox} 
\end{figure}

\clearpage

\begin{figure}[h!]
    \centering
    \begin{tcolorbox}[title={Mathematical Reasoning}]\footnotesize

\texttt{<QUESTION>}

\noindent First, briefly state your step-by-step reasoning. Then, state your final answer in $\backslash\backslash$boxed\{\{\}\} at the very end of your response, just like: ``final answer: $\backslash\backslash$boxed\{\{123\}\}".

    \end{tcolorbox} 
\end{figure}

\subsection{Baseline Prompts}
\vspace{2mm}

\begin{figure}[h!]
    \centering
    \begin{tcolorbox}[title={Self Refinement}]\footnotesize
Previous response:
    
\texttt{<OUTPUT FROM LAST ITERATION>}

Question:

\texttt{<TASK>}

Instructions: Review your previous response above and provide an improved, refined version. Consider:

    1. Accuracy and correctness
    
    2. Clarity and completeness
    
    3. Logical reasoning
    
    4. Better explanations or solutions

Provide your refined response. First, briefly state your step-by-step reasoning. Then, 

\texttt{<TASK-SPECIFIC INSTRUCTIONS>} 

    \end{tcolorbox} 
\end{figure}

\begin{figure}[h!]
    \centering
    \begin{tcolorbox}[title={LLM Judge}]\footnotesize
  
Question:

\texttt{<TASK>}

\noindent Below are \texttt{N} different responses from different agents:

Response 1 (from Agent 1)\\
\texttt{<RESPONSE 1>}\\

Response 2 (from Agent 2)\\
\texttt{<RESPONSE 2>}\\

$\vdots$ \\

Response $N$ (from Agent $N$)\\
\texttt{<RESPONSE N>} \\

\noindent Instructions: Review all the responses above and select the BEST response based on:

    1. Accuracy and correctness
    
    2. Clarity and completeness
    
    3. Quality of reasoning
    
    4. Overall quality

\noindent Your response should be ONLY the number (1, 2, 3, etc.) corresponding to the best response. For example, if you think Response 2 is the best, respond with just ``2".

    \end{tcolorbox} 
\end{figure}

\section{Why the Second Eigenvector of the Laplacian Acts as a Clusterer?}
\label{apdx:fiedler}
Let $G = (V, E)$ be a graph with adjacency matrix $A$ and degree matrix $D$, and define the unnormalized Laplacian $L = D - A$.
For any real-valued function $x \in \mathbb{R}^{|V|}$ defined over the vertices, the quadratic form of the Laplacian is
\[
    x^\top L x = \frac{1}{2}\sum_{i,j} a_{ij}(x_i - x_j)^2.
\]
This quantity measures the smoothness of $x$ over the graph: it is small when adjacent nodes $(i,j)$ have similar values of $x_i$ and $x_j$.
Thus, minimizing $x^\top L x$ encourages $x$ to vary smoothly along edges.

Since $L$ is positive semidefinite, its eigenvalues satisfy
\[
    0 = \lambda_1 \le \lambda_2 \le \cdots \le \lambda_n,
\]
with the first eigenvector $\mathbf{u}_1 = \mathbf{1}$~(one vector) corresponding to the trivial case of no variation across the graph.
The second smallest eigenvector, known as the \emph{Fiedler vector} $\mathbf{u}_2$, solves the constrained optimization problem
\[
    \min_{x \perp \mathbf{1}, \, \|x\| = 1} x^\top L x.
\]
It represents the smoothest nontrivial variation over the graph---that is, the direction along which the graph can be most naturally divided into two weakly connected components.
Nodes with similar $\mathbf{u}_2$ values are strongly connected, whereas nodes with dissimilar $\mathbf{u}_2$ values are weakly connected.
Partitioning the graph by thresholding $\mathbf{u}_2$ (e.g., by its median or sign) therefore yields two clusters that approximately minimize the \emph{graph cut} objective, effectively acting as a binary graph clusterer.

\begin{table*}[h!]
\caption{\textbf{Adjacency matrix similarity function comparison.} We compare our $n$-gram based design choice with embedding cosine similarity-based computation.}
\label{tbl:adjacency}
\setlength{\tabcolsep}{2pt} 
\resizebox{\linewidth}{!}{%
\begin{tabular}{lcccc|cc|c}
\toprule
             & \multicolumn{4}{c}{\textbf{Text Summarization}}   & \multicolumn{2}{c}{\textbf{Code Generation}} & \textbf{Math Reasoning}    \\
\textbf{Method}           & {Rouge-1} & {Rouge-2} & {Rouge-L} & {BLEU} & {Pass@1}   & {BLEU}   & {Accuracy} \\
\midrule
Single Path      &    32.95 $\pm$ 0.36     &   10.47 $\pm$ 0.22     &  20.17 $\pm$ 0.28     &   3.37 $\pm$ 0.18    &   69.89 $\pm$ 3.59  &  7.92 $\pm$ 0.50    &   70.98 $\pm$ 1.74   \\
 \cellcolor{blue!3} ModeX---LastTokenEmb ($N$=16) &  \cellcolor{blue!3}  33.17  &   \cellcolor{blue!3}  10.25 &   \cellcolor{blue!3} 20.26   &    \cellcolor{blue!3} 2.97  &    \cellcolor{blue!3} 75.00   &  \cellcolor{blue!3} 8.20   &   \cellcolor{blue!3} 71.33   \\
 \cellcolor{blue!3} ModeX---SentenceBERT ($N$=16) &  \cellcolor{blue!3}  33.82  &   \cellcolor{blue!3}  \textbf{11.44} &   \cellcolor{blue!3} 20.92   &    \cellcolor{blue!3} 3.77  &    \cellcolor{blue!3} 72.56   &  \cellcolor{blue!3} \textbf{8.53}   &   \cellcolor{blue!3} 70.67   \\
 \cellcolor{blue!3} ModeX---$n$-gram ($N$=16) &  \cellcolor{blue!3}  \textbf{34.28}  &   \cellcolor{blue!3}  11.24 &   \cellcolor{blue!3} \textbf{21.06}   &    \cellcolor{blue!3} \textbf{3.92}  &    \cellcolor{blue!3} \textbf{75.61}   &  \cellcolor{blue!3} 8.45   &   \cellcolor{blue!3} {\textbf{78.00}}   \\
\bottomrule
\end{tabular}
}
\end{table*}
\section{Proofs}
\label{apdx:proofs}

\textit{Proof of Theorem 1.}
Let $\mathcal{G}$ be the similarity graph constructed from samples $V$. The objective of the spectral cut is to find a partition $(V_1, V_2)$ that minimizes the probability flow defined by the conductance $\phi$:
\begin{equation}
    \phi(V_1, V_2) = \frac{\text{cut}(V_1, V_2)}{\min(\text{vol}(V_1), \text{vol}(V_2))}
\end{equation}
where $\text{cut}(V_1, V_2) = \sum_{u \in V_1, v \in V_2} A_{uv}$.
In the limit of large $N$, the graph Laplacian converges to the Laplace-Beltrami operator on the underlying data manifold. 
The Cheeger's Inequality states that the second smallest eigenvalue $\lambda_2$ (associated with the Fiedler vector) bounds the conductance:
\begin{equation}
    \frac{\lambda_2}{2} \le \phi^* \le \sqrt{2\lambda_2}
\end{equation}
If the distribution has two distinct modes separated by a ``valley'' of low probability (low similarity), the edges bridging these regions will have low weights ($A_{uv} \to 0$). This creates a ``bottleneck,'' resulting in a near-zero conductance $\phi^*$.
Consequently, the Fiedler vector cut will optimally slice through this low-density valley, isolating the high-density clusters $\mathcal{M}_1$ and $\mathcal{M}_2$. This ensures that subsequent mode estimation (Theorem 1) is performed \textit{within} a single coherent semantic cluster, preventing the selection of an incoherent ``average'' that lies in the low-probability valley between modes.

\vspace{2mm}
\noindent\textit{Proof of Theorem 2.}
The Kernel Density Estimator $\hat{p}(x)$ for a distribution $p(x)$ given samples $V' = \{v_j\}_{j=1}^N$ is defined as:
\begin{equation}
    \hat{p}(x) = \frac{1}{N h} \sum_{j=1}^N K\left(\frac{x - v_j}{h}\right)
\end{equation}
where $h$ is a bandwidth parameter and $K$ is the kernel.
In our graphical formulation, the edge weight $A_{ij}$ is defined by the similarity $S(v_i, v_j)$, which is the Jaccard similarity measure. Assuming $S$ behaves as a kernel function (where $S(v_i, v_j) \approx K(v_i, v_j)$), the weighted degree of a node $v_i$ is:
\begin{equation}
    d(v_i) = \sum_{j=1}^N A_{ij} = \sum_{j=1}^N S(v_i, v_j)
\end{equation}
Multiplying and dividing by the normalization constants, we observe:
\begin{equation}
    d(v_i) \propto \frac{1}{N} \sum_{j=1}^N K(v_i, v_j) \approx \hat{p}(v_i)
\end{equation}
Thus, the weighted degree $d(v_i)$ is a direct proxy for the local probability density around $v_i$.
\begin{equation}
    v_{\text{centroid}} = \operatorname*{arg\,max}_{v_i \in V'} d(v_i) \equiv \operatorname*{arg\,max}_{v_i \in V'} \hat{p}(v_i)
\end{equation}
Since $V'$ represents a coherent (unimodal) cluster, the sample with the maximum empirical density $\hat{p}(v_i)$ is the consistent estimator for the mode of that cluster.

\section{Similarity Function Comparison for Adjacency Matrix Construction}
\label{apdx:adjacency}

In Table~\ref{tbl:adjacency}, we compare our $n$-gram-based similarity matrix construction with the embedding cosine similarity-based approach.
Specifically, we either retrieve the last token embedding of each output~(LastTokenEmb) or retrieve the sentence embedding using Sentence BERT~\cite{reimers2019sentence}~(SentenceBERT), and compute the cosine similarity between the generated samples:
\begin{equation}
    A_{i,j} = \frac{e_i \cdot e_j}{||e_i|| \times ||e_j||},
\end{equation}
where $e_i, e_j$ refers to the retrieved embeddings for sample $i$ and $j$.
Overall, the both embedding-based methods outperform the Single Path baseline, but is generally worse than ModeX--$n$-gram.
Between the two embedding methods, SentenceBERT performed better on the text summarization task~(CNN/DailyMail), while LastTokenEmb showed stronger results on code generation~(HumanEval) and math reasoning~(Math500). 
We conjecture that this pattern arises because SentenceBERT is better suited for plain text semantics, compared to structural signals present in code and math.

Additionally, we would like to emphasize that we do not claim universal superiority of lexical similarity to embedding-based similarity approaches. 
We fully admit that embedding-based similarity may outperform lexical methods in certain settings with high lexical diversity--especially as stronger embedding models become available. 
In such cases, ModeX can readily incorporate embedding-based similarity in place of Jaccard indices without changing the overall framework.

\begin{table*}[t!]
\centering
\small
\begin{tabular}{lcccc}
\toprule
\textbf{ModeX (N=8) — CNN/DailyMail} & \textbf{ROUGE-1} & \textbf{ROUGE-2} & \textbf{ROUGE-L} & \textbf{BLEU} \\
\midrule
Unigram + Bigram + Trigram & 34.26 & 11.39 & 21.08 & 3.59 \\
\midrule
(-) Unigram & 33.94 & 11.23 & 20.91 & 3.86 \\
(-) Bigram & 33.84 & 11.21 & 20.78 & 3.71 \\
(-) Trigram & 33.77 & 11.19 & 20.78 & 3.70 \\
\bottomrule
\end{tabular}
\caption{Ablation study on n-gram components for similarity construction in ModeX. Each row removes one component from the full model.}
\end{table*}

\section{Ablation on N-gram Components}
\label{apdx:ngram}
We conduct an ablation study to analyze the contribution of different n-gram components (unigram, bigram, and trigram) in constructing the similarity graph used by ModeX. 
Specifically, we evaluate performance on the CNN/DailyMail summarization task by removing one component at a time while keeping the others fixed.

Across all ablations, we observe consistent performance degradation compared to the full model, indicating that each n-gram component contributes meaningfully to the overall performance. Notably, removing trigrams results in the largest drop in performance, while removing unigrams leads to the smallest decrease. This trend aligns with the intuition that higher-order n-grams capture richer semantic and structural information, which is crucial for accurately modeling similarity among generated candidates. In contrast, unigrams provide more coarse-grained lexical signals and therefore have a relatively smaller impact.

\section{Experiments on More Capable Models.}

We extend our evaluation to larger and more capable models, including Qwen2.5-14B and Qwen2.5-32B, across a diverse set of benchmarks spanning summarization, code generation, and mathematical reasoning. 
On the CNN/DailyMail summarization benchmark, ModeX improves over the single-agent baselines for both model sizes. 

\begin{table}[h]
\centering
\small
\setlength{\tabcolsep}{1pt} 
\resizebox{\linewidth}{!}{%
\begin{tabular}{lcccc}
\toprule
\textbf{CNN/DailyMail} & \textbf{ROUGE-1} & \textbf{ROUGE-2} & \textbf{ROUGE-L} & \textbf{BLEU} \\
\midrule
Qwen2.5-14B & $32.34 \pm 0.25$ & $9.67 \pm 0.23$ & $19.90 \pm 0.20$ & $3.04 \pm 0.12$ \\
Qwen2.5-14B + ModeX (N=8) & $33.63$ & $10.37$ & $20.66$ & $3.41$ \\
\midrule
Qwen2.5-32B & $32.84 \pm 0.23$ & $10.18 \pm 0.12$ & $20.03 \pm 0.21$ & $3.19 \pm 0.10$ \\
Qwen2.5-32B + ModeX (N=8) & $34.03$ & $10.88$ & $20.87$ & $3.75$ \\
\bottomrule
\end{tabular}
}
\caption{Results on CNN/DailyMail summarization.}
\end{table}

On the HumanEval benchmark, ModeX substantially improves functional correctness.

\begin{table}[h]
\centering
\small
\setlength{\tabcolsep}{8pt} 
\resizebox{\linewidth}{!}{%
\begin{tabular}{lcc}
\toprule
\textbf{HumanEval} & \textbf{Pass@1} & \textbf{BLEU} \\
\midrule
Qwen2.5-14B & $30.41 \pm 13.25$ & $3.43 \pm 0.53$ \\
Qwen2.5-14B + ModeX (N=8) & $39.02$ & $3.84$ \\
\midrule
Qwen2.5-32B & $28.28 \pm 18.11$ & $3.42 \pm 0.23$ \\
Qwen2.5-32B + ModeX (N=8) & $35.37$ & $3.94$ \\
\bottomrule
\end{tabular}
}
\caption{Results on HumanEval code generation.}
\end{table}

On Math500, ModeX also yields consistent gains in accuracy.

\begin{table}[h]
\centering
\small
\begin{tabular}{lc}
\toprule
\textbf{Math500} & \textbf{Accuracy} \\
\midrule
Qwen2.5-14B & $72.62 \pm 0.92$ \\
Qwen2.5-14B + ModeX (N=8) & $77.67$ \\
\midrule
Qwen2.5-32B & $76.17 \pm 1.32$ \\
Qwen2.5-32B + ModeX (N=8) & $78.67$ \\
\bottomrule
\end{tabular}
\caption{Results on Math500.}
\end{table}

We further evaluate on the AIME2025 benchmark using GPT-4, which consists of 30 challenging mathematical problems. ModeX again demonstrates strong improvements: accuracy increases from $20.42 \pm 2.00$ to $26.67$ with $N=8$, and further to $30.00$ with $N=16$.

\begin{table}[h]
\centering
\small
\begin{tabular}{lc}
\toprule
\textbf{AIME2025} & \textbf{Accuracy} \\
\midrule
GPT-4 & $20.42 \pm 2.00$ \\
GPT-4 + ModeX (N=8) & $26.67$ \\
GPT-4 + ModeX (N=16) & $30.00$ \\
\bottomrule
\end{tabular}
\caption{Results on AIME2025.}
\end{table}

Overall, these results show that ModeX consistently outperforms single-agent baselines even as model capability scales, indicating that its benefits are not limited to smaller models but extend to more advanced systems.
\onecolumn
\clearpage
\section{Algorithms}
\label{apdx:algo}

\begin{algorithm*}[h!]
\caption{Mode Extraction~(ModeX)}
\label{alg:mode_selection}
\begin{algorithmic}[1]
\Require LLM $\mathcal{F}$; Number of text paths $N$; Similarity function $\mathrm{Sim}(\cdot,\cdot)$; Spectral clustering routine $\mathrm{SpecCluster}(\cdot)$; Cluster evaluator $\mathrm{CutCriterion}(\cdot,\cdot)$; Spectral threshold $\tau$.
\Ensure Selected response $r_{i^*}$
\State \textbf{Initialize} active index set $S \leftarrow \{1, 2, \dots, n\}$.
\State \textbf{Initialize} response set $R \gets \{r_i\}_{i\in S}, \quad$ where $r_i \leftarrow \mathcal{F}(\text{prompt}_i)$
\State \textbf{Initialize}  $A,\quad$ where $A_{ij} \leftarrow \mathrm{Sim}(r_i, r_j) \qquad \forall\; r_i, r_j \in R$ \Comment{(1) adjacency matrix construction}
\While{$|S| > 1$}
    \State $(C_1, C_2) \leftarrow \mathrm{SpecCluster}(A)$ \Comment{(2) graph spectral clustering}
    \If{CutCriterion$(C_1, C_2) < \tau$}
        \If{$|C_1| \neq |C_2|$}
            \State $S \gets \arg\max_{C \in \{C_1,C_2\}} \; |C|$
        \Else
            \State $S \gets \arg\max_{C \in \{C_1,C_2\}} \; \sum_{i\in C, j\in[N]} A_{ij}$
        \EndIf
        \State Update $A, R$ to include only $i,j \in S$
    \Else
        \State Terminate clustering
    \EndIf
\EndWhile
\State $i^* \leftarrow \text{maximum degree node index of $A$}$ \Comment{(3) centroid selection}
\State \Return $r_{i^*}$
\end{algorithmic}
\end{algorithm*}

\begin{algorithm*}[h!]
\caption{ModeX-Lite}
\label{alg:mupath}
\begin{algorithmic}[1]
\Require LLM $\mathcal{F}$; Input prompt $x$; Initial number of paths $N$; Pruning interval $T$
\Ensure Final response $y$
\State Initialize $N$ generation paths: $X^{(0)} \gets \{x^{(i)} = x\}_{i=1}^{N}$
\State $t \gets 0$
\While{not all remaining paths have generated an \texttt{EOS} token}
    \State $X^{(t+1)} \gets \mathcal{F}\!\left(X^{(t)}\right)$ \Comment{one-step parallel generation}
    \If{$(t+1) \bmod T = 0$}
        \State $X^{(t+1)} \gets \textsc{ClusterAndPrune}\!\left(X^{(t+1)}\right)$
        \Comment{adjacency construction \& spectral clustering}
    \EndIf
    \State $t \gets t + 1$
\EndWhile
\State $y \gets \textsc{SelectCentroid}\!\left(X^{(t)}\right)$ \Comment{centroid selection}
\State \Return $y$
\end{algorithmic}
\end{algorithm*}

\clearpage
\section{More Qualitative Examples}
\label{apdx:more_qual}

\begin{figure*}[th!]
\begin{tcolorbox}[title={Chosen vs. Rejected}~(CNN/DailyMail)]
\begin{lstlisting}
(*\textbf{\color{blue}{[Chosen]}}*) The article discusses Tiger Woods' performance at the Masters, focusing on his struggle with the short game while showing improvement in driving. Despite his challenges, particularly with his chipping, the galleries remained supportive. Woods showcased moments of his former greatness but was far from his peak performance, dealing with an ongoing crisis of confidence and physical ailments. His game demonstrated both his cussedness and vulnerability, indicating mixed progress towards regaining his competitive form.

(*\textbf{\color{magenta}{[Rejected]}}*) Tiger Woods experienced a challenging start to his first round at the Masters, showcasing both flashes of brilliance and persistent flaws. Despite working on his short game, Woods exhibited inconsistent driving, which overshadowed his typically strong chipping skills. His performance highlighted a mix of improvement and decline, with Woods making heroic saves and committing occasional errors. A notable moment included his stunning drive on the seventh hole, reminiscent of his past greatness, juxtaposed with misses like a mis-hit approach to the tenth hole. The crowd, though supportive, recognized the signs of an older Tiger Woods. While showing promise and determination, Woods' performance suggested a player still adjusting to physical limitations and potentially dealing with psychological challenges.

\end{lstlisting}
\end{tcolorbox}
\end{figure*}

\begin{figure*}[th!]
\begin{tcolorbox}[title={Chosen vs. Rejected}~(CNN/DailyMail)~(CNN/DailyMail)]
\begin{lstlisting}
(*\textbf{\color{blue}{[Chosen]}}*) George Boyd is the Barclays Premier League's most tireless player, having covered 210.5 miles this season, a feat equivalent to running between Burnley's Turf Moor and Crystal Palace's Selhurst Park. Boyd beats fellow player Christian Eriksen in the "running marathon." Burnley, sitting near the relegation zone, boasts three players in the top 10 hardest-working runners, collectively covering 2,172.3 miles.

(*\textbf{\color{magenta}{[Rejected]}}*) George Boyd of Burnley has set a new milestone as the hardest working player in the Barclays Premier League this season, having run 215 miles. This distance equals the journey from Turf Moor to Selhurst Park, with Boyd leading the list just ahead of Tottenham's Christian Eriksen. Burnley, currently 18th and fighting for survival, has three players in the top 10 running enthusiasts. Key other featured players include Stewart NZonzi, Jordan Henderson, and Daryl Janmaat. Boyd, Arfield, and Jones lead a total running distance of 2,172.3 miles for Burnley. The ranking highlights efforts from lower-ranked teams, with only a few from top-half clubs making the list. Boyd aims to use his running for Burnley's benefit in upcoming matches against Tottenham.

(*\textbf{\color{black}{[Target]}}*) Burnley winger George Boyd has run the furthest in the Premier League. His 210.5 miles this season beats Christian Eriksen from Tottenham. Burnley have run an incredible 2,172.3 miles between them this season. Players from Stoke, Hull, Liverpool and West Ham feature in the top 10.

\end{lstlisting}
\end{tcolorbox}
\end{figure*}

\begin{figure*}[th!]
\begin{tcolorbox}[title={Chosen vs. Rejected}~(CNN/DailyMail)]
\begin{lstlisting}
(*\textbf{\color{blue}{[Chosen]}}*) The article discusses the rapid growth in the number of millionaires in various African cities, particularly in Nigeria's Lagos and others like Luanda, Dar es Salaam, and Accra. This new affluent class is driving the demand for luxury goods, with sales of high-end products growing significantly in these regions. Cities like Nigeria are now rivaling traditional wealth centers for luxury consumption. Market experts attribute this growth to rising incomes and population. These wealthy consumers, often younger than their Western counterparts, prefer well-known global brands that symbolize status and success. Meanwhile, some homegrown African businesses see this trend as an opportunity to establish themselves in the luxury market, leveraging the continent's rich cultural heritage and craftsmanship.

(*\textbf{\color{magenta}{[Rejected]}}*) The number of millionaires in African cities like Lagos, Luanda, Dar es Salaam, and Accra is rapidly increasing, with significant growth particularly noted in Lagos. This shift has transformed the consumption patterns of luxury goods, with wealthy Nigerians now able to enjoy prestigious brands locally. Factors such as rising incomes and slowing growth in other emerging markets are driving this trend. While international luxury brands remain dominant, there is a growing interest in local, artisanal products as Africa's affluent consumers become younger and look to luxury for social status. Despite the promising outlook, the future of luxury markets in Africa depends on economic reforms improving infrastructure and trading conditions. Additionally, there is a recognition of the potential for developing a strong domestic luxury industry by leveraging traditional craftsmanship and cultural heritage.

(*\textbf{\color{black}{[Target]}}*) The city with most multimillionaires in Africa is Johannesburg. However a crop of new pretenders have been expanding their millionaire count.
\end{lstlisting}
\end{tcolorbox}
\end{figure*}

\begin{figure*}[th!]
\begin{tcolorbox}[title={Chosen vs. Rejected}~(CNN/DailyMail)]
\begin{lstlisting}
(*\textbf{\color{blue}{[Chosen]}}*) A 16-year-old student, Rebecca Rice, was allowed to bring her pet Labrador, Holly, into a GCSE French oral exam to help calm her nerves. This arrangement was made after Rebecca's teacher, Lorette Esteve, learned that no other methods were effective. Bodmin College in Cornwall, where the student attends, has implemented various strategies to reduce exam stress, such as bringing drinks, lucky items, or friends into the exam room.


(*\textbf{\color{magenta}{[Rejected]}}*) A 16-year-old student, Rebecca Rice, was allowed to bring her 11-year-old Labrador, Holly, into a French GCSE oral exam to help manage her severe nerves. Following a discussion with her teacher, Lorette Esteve, who leads foreign languages at Bodmin College, it was decided that Holly's presence would assist Rebecca, as nothing else did. Rebecca felt her nerves settled significantly with Holly by her side during the five-minute exam. Esteve mentioned that the school employs various methods, such as offering drinks, lucky key rings, or friends in the room, to help students cope during exams.

(*\textbf{\color{black}{[Target]}}*) Rebecca Rice, 16, was allowed to take her pet dog into five-minute exam. She told her teacher she struggles to control her nerves ahead of the GCSE. Teacher Lorette Esteve advised she bring along 11-year-old dog Holly. School has various tactics help reduce stress including lucky key rings.
\end{lstlisting}
\end{tcolorbox}
\end{figure*}

\begin{figure*}[th!]
\begin{tcolorbox}[title={Chosen vs. Rejected}~(CNN/DailyMail)]
\begin{lstlisting}
(*\textbf{\color{blue}{[Chosen]}}*) Dr. Xiao-Ping Zhai, based in Harley Street, London, uses traditional Chinese medicine to help women conceive. She offers methods like acupuncture and customized Chinese herb prescriptions, claiming to have helped over a thousand women get pregnant. Patients like Jane Parker successfully conceived using Dr. Zhai's techniques. Her methods are detailed in a BBC documentary and include daily herbal tea or vitamin tablets, costing up to (*\pounds*)150 weekly. The program also explores other alternative treatments, including leech therapy, reflecting a trend towards traditional remedies among infertile couples unsatisfied with Western medicine.

(*\textbf{\color{magenta}{[Rejected]}}*) Dr. Xiao-Ping Zhai, a prominent figure at The Zhai Clinic in London, uses traditional Chinese medicine to help infertile couples conceive. Opened in 1996, her clinic offers services such as acupuncture and customized Chinese herbal remedies, which she claims have helped over a thousand women become pregnant. The BBC documentary "Inside Harley Street" provides insight into Zhai's approach and features the journey of patient Jane Parker, who conceived her son at age 41 through the clinic's methods. While many seek these alternative treatments due to dissatisfaction with conventional approaches like IVF, other practices, such as leech therapy, are also explored for their purported benefits in various health issues. The documentary examines the growing trend of relying on non-traditional methods for overcoming infertility, reflecting the desire among patients for personalized and holistic care.

(*\textbf{\color{black}{[Target]}}*) Around one in seven couples suffer with infertility. Dr Xiao-Ping Zhai offers help via with traditional Chinese medicine. Uses acupuncture and prescribes course of Chinese herbs.
\end{lstlisting}
\end{tcolorbox}
\end{figure*}

\begin{figure*}[th!]
\begin{tcolorbox}[title={Chosen vs. Rejected}~(CNN/DailyMail)]
\begin{lstlisting}
(*\textbf{\color{blue}{[Chosen]}}*) Australian cook Lyndey Milan collaborated with Aldi to create affordable Easter recipes that cost less than $6 per person. She developed meals like Greek lamb with salad and zucchini pilaf, herb-crusted salmon, and desserts such as hot cross bun and Easter egg bread & butter pudding. Milan stressed the cost-effectiveness, health benefits, and fun aspects of homemade cooking. Practical tips include using up all bought ingredients, purchasing seasonal produce, and freezing surplus foods to reduce waste and grocery expenses.

(*\textbf{\color{magenta}{[Rejected]}}*) Lyndey Milan, an Australian chef known for her home cooking expertise, has collaborated with Aldi to create affordable Easter lunch recipes that cater to six people for under $6 per person. Utilizing fresh, seasonal produce from Aldi, Milan offers recipes like Greek lamb with salad and zucchini pilaf, herb-crusted salmon with pea puree, smashed potatoes, and carrots, along with a hot cross bun and Easter egg bread and butter pudding. She emphasizes tips for minimizing waste, reducing grocery expenses, and maximizing the use of fresh ingredients, suggesting that her dishes are not only budget-friendly but also tasty and easy to prepare.

(*\textbf{\color{black}{[Target]}}*) Lyndey Milan created Easter lunch recipes for $6 per person from Aldi. Home cook icon trawled supermarket aisles to find inspiration for recipes. Result of experiment include Greek lamb and herb crusted salmon dishes .
Hot cross bun and Easter egg bread & butter pudding came in at just $2 a person and Lyndey says it is sure to be a crowd pleaser.

\end{lstlisting}
\end{tcolorbox}
\end{figure*}

\begin{figure*}[th!]
\begin{tcolorbox}[title={Chosen vs. Rejected}~(HumanEval)]
\begin{lstlisting}
(*\textbf{\color{black}{[Task]}}*) 
Complete the following code:


def specialFilter(nums):
    """Write a function that takes an array of numbers as input and returns 
    the number of elements in the array that are greater than 10 and both 
    first and last digits of a number are odd (1, 3, 5, 7, 9).
    For example:
    specialFilter([15, -73, 14, -15]) => 1 
    specialFilter([33, -2, -3, 45, 21, 109]) => 2
    """


 Make sure to provide ONLY your COMPLETE code after stating "# Code # ".
 
(*\textbf{\color{blue}{[Chosen]}}*) 
def specialFilter(nums):
    def is_odd_digit(n):
        return n % 2 != 0
    
    def has_odd_first_last_digits(num):
        str_num = str(abs(num))
        first_digit = int(str_num[0])
        last_digit = int(str_num[-1])
        return is_odd_digit(first_digit) and is_odd_digit(last_digit)
    
    count = 0
    for num in nums:
        if num > 10 and has_odd_first_last_digits(num):
            count += 1
    return count

(*\textbf{\color{magenta}{[Rejected]}}*) 
def specialFilter(nums):
    def is_odd_digit(n):
        while n > 0:
            digit = n % 10
            if digit not in [1, 3, 5, 7, 9]:
                return False
            n //= 10
        return True
    
    count = 0
    for num in nums:
        if num > 10 and is_odd_digit(abs(num)):
            count += 1
            
    return count
\end{lstlisting}
\end{tcolorbox}
\end{figure*}

\begin{figure*}[th!]
\begin{tcolorbox}[title={Chosen vs. Rejected}~(HumanEval)]
\begin{lstlisting}
(*\textbf{\color{black}{[Task]}}*) 
Complete the following code:

def f(n):
    """ Implement the function f that takes n as a parameter,
    and returns a list of size n, such that the value of the element at index i is the factorial of i if i is even
    or the sum of numbers from 1 to i otherwise.
    i starts from 1.
    the factorial of i is the multiplication of the numbers from 1 to i (1 * 2 * ... * i).
    Example:
    f(5) == [1, 2, 6, 24, 15]
    """

 Make sure to provide ONLY your COMPLETE code after stating "# Code # ".
 
(*\textbf{\color{blue}{[Chosen]}}*) 
def f(n):
    def factorial(x):
        if x == 0 or x == 1:
            return 1
        else:
            return x * factorial(x - 1)
    
    def sum_to_i(x):
        return x * (x + 1) // 2
    
    result = []
    for i in range(1, n + 1):
        if i % 2 == 0:
            result.append(factorial(i))
        else:
            result.append(sum_to_i(i))
    return result

(*\textbf{\color{magenta}{[Rejected]}}*) 
import math
def f(n):
    def factorial(i):
        return math.factorial(i)
    
    def sum_to_i(i):
        return i * (i + 1) // 2
    
    result = []
    for i in range(1, n + 1):
        if i % 2 == 0:
            result.append(factorial(i))
        else:
            result.append(sum_to_i(i))
    return result
\end{lstlisting}
\end{tcolorbox}
\end{figure*}

\end{document}